\pdfoutput=1

\documentclass[11pt]{article}

\usepackage{acl}
\usepackage{graphicx}
\usepackage{times}
\usepackage{latexsym}
\usepackage[T1]{fontenc}
\usepackage[utf8]{inputenc}
\usepackage{microtype}
\usepackage{multirow}
\def\wtkappa{quadratic weighted kappa}

\title{Short Answer Grading Using One-shot Prompting and Text Similarity Scoring Model}

\author{Su-Youn Yoon \\
  EduLab, Inc., \\
  Shibuya city, Tokyo, Japan \\
  \texttt{su-youn.yoon@edulab-inc.com} \\}

\begin{document}
\def\answerspans{justification keys}
\def\answerspan{justification key}
\maketitle

\begin{abstract}
In this study, we developed an automated short answer grading (ASAG) model that provided both analytic scores and final holistic scores. Short answer items typically consist of multiple sub-questions, and providing an analytic score and the text span relevant to each sub-question can increase the interpretability of the automated scores. Furthermore, they can be used to generate actionable feedback for students. Despite these advantages, most studies have focused on predicting only holistic scores due to the difficulty in constructing dataset with manual annotations. To address this difficulty, we used large language model (LLM)-based one-shot prompting and a text similarity scoring model with domain adaptation using small manually annotated dataset. The accuracy and quadratic weighted kappa of our model were $0.67$ and $0.71$ on a subset of the publicly available ASAG dataset. The model achieved a substantial improvement over the majority baseline.
\end{abstract}

\section{Introduction}
Automated scoring systems can assess learners' responses faster than human raters, with the resulting scores being consistent over time. This has prompted strong demand for high-performing automated scoring systems.

Items that elicit short responses are efficient method to evaluate students' knowledge and have been widely used in the education field. Many researches have been conducted in developing automated short answer grading (ASAG) systems resulting high performing systems with relatively small training dataset. 

\begin{figure*}[htbp]
  \centering
  \includegraphics[scale=0.45]{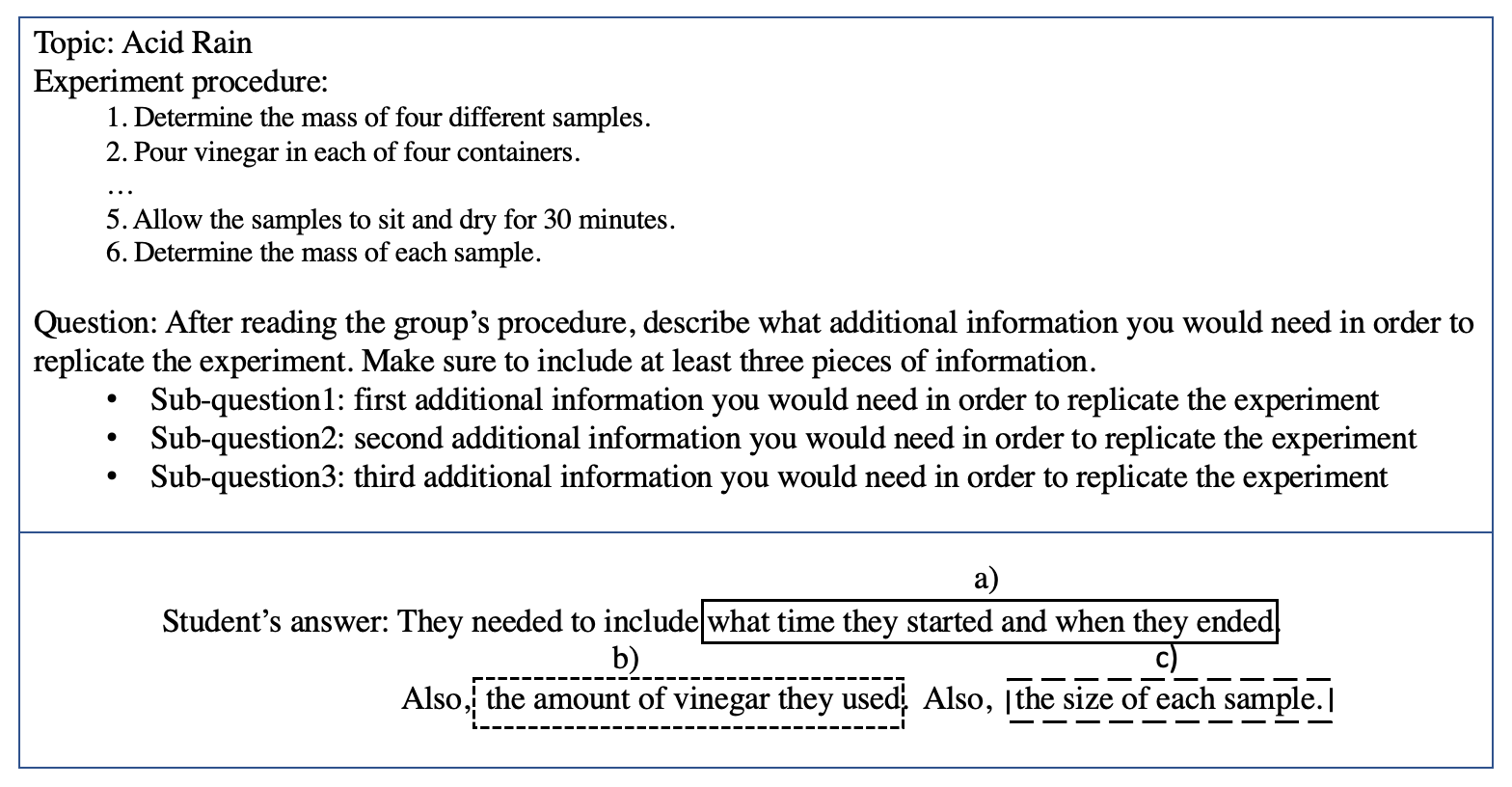}
  \caption{Example of a question and student's answer from the Automated Student Assessment Prize Short Answer Scoring dataset}
\label{fig1}
\end{figure*}

Short response grading focuses on the content rather than the writing quality, and the possible answers are restricted by the closed-form of the question \cite{burrows2015}. Typically, the scoring rubric for one question consists of multiple sub-questions that are in the form of "if an answer includes content X, then it receives a score of Y". Figure \ref{fig1} provides an example of the question and student's answer from the publicly available ASAG dataset. The question requests answers for three sub-questions, and the relevant text spans of each answer is marked by (a), (b), (c) in Figure \ref{fig1}. We will refer to the text span relevant to each sub-question as the `justification key' and the score of each sub-question as the `analytic score' as in \newcite{mizumoto2019}.

Providing analytic scores and justification keys is beneficial because it can improve the interpretability and validity of automated scoring models. For instance, by comparing the system-detected justification keys with the manually-annotated justification keys, we can understand where the systems pay attention during score generation and evaluate whether it is valid or not. Next, they can be used for systems to provide actionable feedback; students can understand why their answers are incorrect and how they can correct them. However, it is difficult to train such models because it requires additional human annotations, resulting in a substantial cost increase.

In this study, we will present an automated short answer grading model that can produce analytic scores and justification keys. To overcome the difficulty of constructing a large scale dataset with manual annotations, we will explore approaches that can be trained using small training data. 

\section{Previous studies}
Previous studies in the ASAG field have explored various approaches, such as traditional linguistic features with statistical models and neural models. Among these possible approaches, the appropriate one in a particular scoring condition is largely dependent on the availability of a manually annotated, question-specific dataset. If sufficient amount of human-scored answers are available for each question as training data, data-driven models such as statistical models using word n-gram features or embedding-based neural models have shown promising performance\cite{heilman2013,riordan2017, sung2019}. On the contrary, if only rubrics or small reference samples are available, the models need to rely on methods to calculate the semantic similarity between the reference and the test answers.
 
Among the first group, \newcite{heilman2013} developed a model based on the word and character n-gram features and domain adaptation approach, and it achieved good performance. \newcite{riordan2017} investigated several basic neural architectures and the model comprised of the word embeddings, LSTM layers, and attention could achieve promising performance over a wide range of tasks and datasets. More recently, \newcite{sung2019} achieved the comparable performance to human raters by fine-tuning pre-training BERT model with domain adaptation.

Most studies have focused on predicting only holistic scores, but some recent studies have investigated predicting analytic scores and \answerspans. \newcite{mizumoto2019} trained a bidirectional LSTM model with an attention mechanism to minimize the difference between the system attention span and manually annotated justification keys. The model achieved good performance in both tasks, and the joint-learning could improve the performance of the holistic score prediction in the low-resource scenario. \newcite{wang2021} used a self-learning approach and replaced the automatically detected \answerspans~with the manual annotations. In the low-resource scenario, the addition of automated annotations had a positive impact, but this impact decreased as the manually-scored data increased.

The researches in the second group consider the ASAG to be a combination of two well-known Natural Language Processing (NLP) tasks: answer-span (justification key) extraction and text-similarity scoring\cite{haller2022}. First, reference answers are created based on the rubrics or model answers. Next, they extract justification keys from a student's answer using regular expressions or methods used in the information retrieval. The similarity scores between student's answers and reference answers were calculated using various semantic similarity metrics such as bag-of-words, content vector analysis, and word embedding. Based on the similarity scores, a final score that decides whether the justification key is semantically equivalent to the concept in the reference answers is assigned. 

Similar to the second group, we aim to solve the ASAG task using only a few manually-scored answers and the two-task approaches. Especially, we will explore LLM-based one-shot prompting in the context of the ASAG. Recent studies\cite{radford2019, brown2020, chung2022} have demonstrated that large language models (LLMs) with larger sizes of parameters can achieve substantial performance improvements over small pre-trained language models. Furthermore, unlike pre-trained language models which require fine-tuning with a substantially large task-specific dataset, LLMs can learn new tasks through in-context learning using only a few demonstration examples. In particular, GPT-3, the third generation of the Generative Pre-trained Transformer model, has shown superior performance in few-shot learning, and it has achieved comparable or superior performance over fine-tuned pre-trained models on diverse tasks\cite{brown2020}.

LLM-based few-shot learning has already been used in educational tasks such as writing assistant, tutoring\cite{cao2023} or item generation\cite{park2022}, but relatively limited researches have conducted in the automated scoring field. \newcite{mizumoto2023} used GPT-3 to grade holistic writing proficiency of non-native English learners' essays, but the performance was not high.

In this study, we will explore a combination of LLM-based one-shot prompting and sentence-similarity scoring model. In particular, we aim to answer the following questions:
\begin{itemize} 
\item How accurately can the model grade short answers? 
\item Can unsupervised domain adaptation achieve substantial improvement over the baseline model without the domain adaptation?
\item How accurately can the model predict the \answerspans~and analytic scores?
\end{itemize}

The proposed model produces both \answerspans~and analytic scores, and it is essential to evaluate the quality of these outputs. However, the data used in this study included only holistic scores. Because of this, we generated holistic scores based on the analytic scores and indirectly evaluated the quality of the analytic scores through the holistic score evaluation. For the \answerspan~evaluation, we used a small set of manually annotated dataset. 

\section{Data}
We used the subset of the Automated Student Assessment Prize Short Answer Scoring (hereafter, ASAP-SAS) data set. It originally included ten questions from science, biology, English, and English language arts domains. We selected four questions (two science and two biology\footnote{They were question 1, 2, 5, and 6.}) because the rubrics included the detailed explanations and the correct responses.  

The total number of answers for the four questions were $6,542$. The data was partitioned into train(80\%), development(10\%), and test(10\%) set, respectively. From the train set, $64$ answers ($16$ answers per question) were selected for the reference answer augmentation (hereafter, answer augmentation set)

\begin{table}[htbp]
  \centering
   \begin{tabular}{|p{0.4cm}|p{0.8cm}|p{1.8cm}|p{1.9cm}|p{0.5cm}|}
\hline
\multirow{2}{*}{} &	\multicolumn{2}{|c|}{Train} & \multirow{2}{*}{Development} & \multirow{ 2}{*}{Test}  \\\cline{2-3}
& Total & Answer augmentation  &  &  \\\hline
N  & 5,226 & 64 & 680  & 636 \\ \hline
    \end{tabular}%
  \caption{Number of answers in each set}
    \label{tab:size}%
\end{table}%

Table \ref{tab:hscore} shows the distribution of the holistic scores (score1) rated by a human rater. The majority score was $0$ and it was more than half of the entire data. 
\begin{table}[htbp]
  \centering
   \begin{tabular}{|c|c|c|c|c|}
\hline
Human score &	0  & 1 & 2 & 3 \\\hline
\%  & 53 & 19 &  17 & 11 \\ \hline
    \end{tabular}%
  \caption{Percentage of each score out of the total data}
    \label{tab:hscore}%
\end{table}%

For each answer in the answer augmentation set, we manually annotated the \answerspans~and analytic scores. All four questions used in this study consisted of three sub-questions. We marked the \answerspans~of each answer as (a), (b), (c) in Figure \ref{fig1}. Next, we assigned a binary score for each \answerspan. The score was 1 when the \answerspan~matched one of the correct answers provided in the rubrics. If the answer received a holistic score of $3$ (perfect score) but the \answerspan~was not included in the correct answers, we still treated it as a correct answer. We assigned $1$ and registered this \answerspan~to the correct answer list\footnote{The correct answers in the rubrics did not cover all possible correct answers. In order to increase the efficiency of finding missing correct answers, we annotated the answers with a holistic score of $3$ and registered missing answers first.}. For the remaining cases, we assigned $0$\footnote{For instance, in Figure \ref{fig1}, the analytic scores of (a), (b), (c) were 0, 1, 0, respectively.}. In addition, we stored the correct answer that matched each \answerspan.

\section{Method}
We used an LLM-based one-shot prompting and a neural text similarity model to grade students' answers. First, we extracted a \answerspan~for each sub-question using the one-shot prompting. Next, we calculated the similarity scores between each \answerspan~and the reference answers using the Sentence-BERT\cite{reimers2019} model and assigned an analytic score. Finally, we generated a holistic score from all analytic scores. A detailed explanation for each step is provided in the subsection.

\subsection{Justification key extraction}
\begin{figure*}[ht]
  \centering
  \includegraphics[scale=0.45]{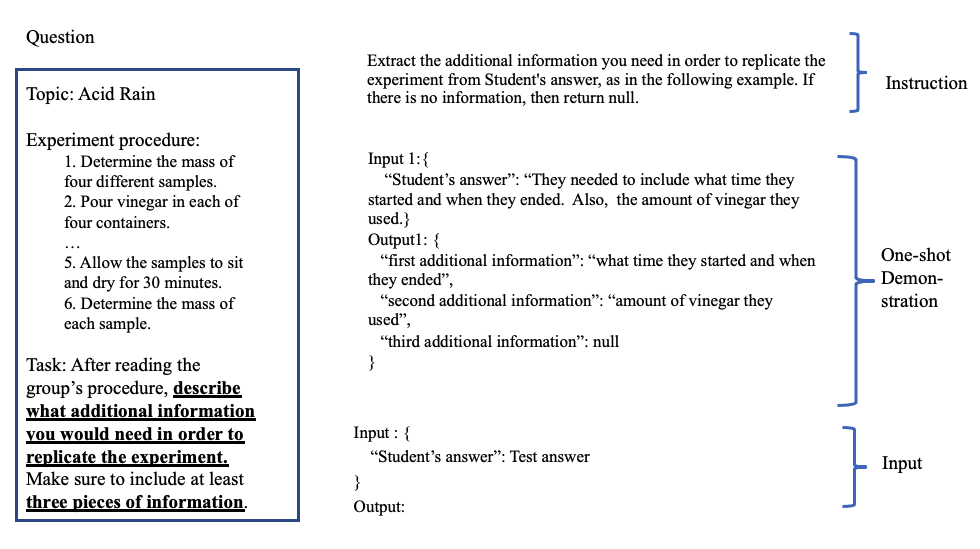}
  \caption{one-shot prompt for the question in Figure 1}
\label{fig2}
\end{figure*}

We used an one-shot prompt to extract \answerspans~from the student's answer. The prompt consisted of the instruction and one demonstration example. An example of one question and its prompt is presented in Figure \ref{fig2}.

We extracted the text describing the sub-questions from the original question (underlined sentences in Figure \ref{fig2}) and used it as an instruction. The demonstration example consisted of an input and an output in JSON format. The input contained a "student's answer" field with one sample answer as its value. The output contained \answerspans~for each sub-question extracted from the sample answer.

\subsection{Reference Answer Augmentation}
We extracted correct answers from the rubrics as reference answers. The initial reference answers were augmented using the manually annotated answer augmentation set. The \answerspans~were extracted and classified into two groups (correct vs. incorrect) based on the manual analytic scores. In addition, for each correct answer in the augmented set, we stored the most similar correct answer from the initial reference answers. We used this information to expedite labeling of the gold dataset used in the domain adaptation. After removing duplicated keys, the \answerspans~were combined with the initial reference answers.   

\subsection{Sentence-BERT and Domain Adaptation}
\label{method:sbert_adaptation}
Sentence-BERT (hereafter, SBERT) is a modification of the pre-trained BERT network, and it uses Siamese network and triplet loss to minimize the distance between sentence pairs with similar meaning s and maximize the distance between sentence pairs with different meanings \cite{reimers2019}. The model derives semantically meaningful sentence embeddings, and the similarity score between two sentence embeddings (e.g., cosine similarity score) is used at the inference time. The models trained on the natural language inference (NLI) dataset have showed the state-of-the art performance in various unsupervised text similarity scoring tasks.  

The performance of SBERT can be further improved by domain adaptation \cite{thakur2020}. For domain adaptation, we created two datasets: a dataset with manual annotations (hereafter, gold dataset) and a dataset with automatically generated labels(hereafter, silver dataset).

For the gold dataset, we used the answer augmentation set. For each student's answer in the set, we extracted \answerspans~and created all possible combinations with the augmented reference answers. The label for each combination was generated in the semi-automated way. If the text pairs of the combination was semantically equivalent, the label was $1$, otherwise it was $0$. We used different steps for the correct answer group and the incorrect answer group. If the \answerspan~was from the correct answer group, we compared its most similar reference answer with that of the reference text of the combination. If they were identical, we assigned $1$; otherwise, we assigned $0$. If the \answerspan~was from the incorrect answer group and the reference text was from the correct answer group, we assigned $0$. For the remaining cases, we manually examined the combination and assigned $1$ when they were semantically equivalent; otherwise, we assigned $0$. 

The silver dataset was generated using the training partition. All answers in the training set were split into sentences. For each sentence, we found two answers from the reference answers: the most similar correct answer and the most similar incorrect answer. 
A similarity score between the training sentence and the reference answer was calculated using the pre-trained SBERT cross-encoder. We first selected one with the lower similarity score and assigned $0$. For the remaining one, if the score was higher than $0.5$,  the label was $1$; otherwise it was $0$. 

Finally, we trained the pre-trained SBERT bi-encoder model using both gold and silver dataset.

\subsection{Analytic Score Generation}
At the inference, \answerspans~in the student's answer were extracted using the prompt. For each \answerspan, we created all possible pairs with the reference answers and calculated the similarity scores using the adapted SBERT model described in the section \ref{method:sbert_adaptation}. The pair with the highest similarity score was selected and the analytic score was generated as follows: the analytic score was 1 only when the most similar reference answer was one of the correct answers and the similarity score was higher than the $0.5$. Otherwise, it was 0.

\subsection{Holistic Score Generation}
Holistic score was the sum of all analytic scores. There was no penalty for incorrect answers. The maximum score for all questions in this study was $3$, so no extra points were awarded for answers containing more than 3 correct \answerspans.

\section{Experiment}
We used the OpenAI’s GPT-3.5 `text-davinci-003' model for the one-shot prompting. We used the text completion API and set the temperature to 0 to avoid the randomness in the output.

For the SBERT model, we used the `all-mpnet-base-v2 model' in the huggingface model repository\footnote{https://huggingface.co/sentence-transformers/all-mpnet-base-v2} as the base model. 

For the domain adaptation, we used both a gold dataset with the manual labels and a silver dataset with the automated labels. The size of the gold dataset was $1,218$ text pairs, while the size of  silver dataset was $33,000$ text pairs. The base SBERT model was trained using the combined dataset. 

As a bench-marking model, we fine-tuned pre-trained language model \cite{devlin2018} using the training script provided by \newcite{zeng2022}\footnote{The training script was downloaded from https://github.com/douglashiwo/AttentionAlignmentASAS}. For each question, the pre-trained bert-base-cased model with a sequence classification layer on top was fine-tuned using the training set. The detailed descriptions about the training was provided in \newcite{zeng2022} (hereafter, fine-tuned BERT model)

The proposed model used human scores for $64$ answers, while the bench-marking model used the human scores for the entire training set($5,226$).

\section{Results}
Table \ref{tab:res} provides the performance of the automated scoring models on the test set. The human raters achieved a good agreement in this task, and both the exact agreement and \wtkappa~were $0.92$ and $0.96$, respectively.

\begin{table}[htbp]
\centering
\begin{tabular}{|p{2.5cm}|c|c|c|}
\hline
	& accuracy	& wtkappa	& corr \\\hline
one-shot+SBERT	& 0.68 &	0.73 &	 0.73 \\\hline
fine-tuned BERT	& 0.77	& 0.88	& 0.88 \\\hline
\end{tabular}
\caption{accuracy, quadratic weighted kappa, and Pearson correlation coefficients between the human and system scores}
\label{tab:res}
\end{table} 

The combination of one-shot prompting and SBERT (one-shot+SBERT) achieved substantially better performance over the majority baseline (accuracy = $0.53$). However, there was a meaningful difference with the fine-tuned BERT model, and the fine-tuned model achieved substantially better performance. 

The one-shot+SBERT model used over $1,000$ non-scored answers per question. In order to evaluate the impact of non-scored answers on the model performance, we conducted an ablation test. While using the same method for the \answerspan~ extraction, we used following additional SBERT models for the semantic similarity scoring:

\begin{itemize}
\item baseline: the baseline SBERT model without the reference answer augmentation
\item reference answer-augmented model: the baseline SBERT model with the reference answer augmentation
\item SBERT with the domain adaptation: the SBERT model adapted using the training set without the reference answer augmentation
\end{itemize}

Table \ref{tab:ablation} provides the results of the ablation test.

\begin{table}[htbp]
\centering
\begin{tabular}{{|p{2.5cm}|c|c|c|}}
\hline
	& accuracy	& wtkappa	& corr \\
\hline
baseline	& 0.44 &	0.48 & 0.46	  \\
\hline
reference answer-augmented model & 0.52 &	0.56 & 0.62	 \\
\hline
SBERT with the domain adaptation	& 0.67	& 0.71	& 0.74  \\
\hline
one-shot+SBERT	& 0.68	& 0.73	& 0.73  \\
\hline
\end{tabular}
\caption{Ablation test of the one-shot+SBERT model}
\label{tab:ablation}
\end{table}

The performance of the baseline model without any adaptation on the ASAG dataset was substantially worse than the final model (one-shot+SBERT). The accuracy was $0.44$ and it was even lower than the majority baseline. The augmentation of the reference answers using a small dataset could achieve a small improvement, but the major improvement was achieved by the domain adaptation. The impact of the reference answer augmentation was marginal when the model was adapted to the domain dataset; differences between adapted SBERT model with and without the answer augmentation were lower than $0.02$ in both accuracy and the quadratic weighted kappa. 

\section{Discussion}
We investigated the accuracy of detecting \answerspans~using the answer augmentation set. We compared manually annotated \answerspans~with those extracted by the GPT-3.5 model. Based on the manual annotation, there were $216$ \answerspans, resulting in an average of $3.37$ \answerspans~per answer. We calculated the the number of words in the \answerspans. The average based on the manual annotations was $8.4$, while that based on the GPT model was $10.4$. The GPT model had a tendency to extract longer spans and contained a few extra words such as conjunction markers at the beginning. Adding these extra words did not have a big impact on the next step, so we treated these cases as correct. More concretely, for each manual and system-based justification key pairs, if the system key contained the entire manual key or the word overlap between two keys was over 90\%, we counted it as a correct case. The correct cases were 195 out of 216, resulting in an accuracy of 90.3\%. The accuracy of the \answerspan~  detection was high. Furthermore, most errors were from low-scoring answers where the justification keys did not contain correct content. These errors did not have an impact on the overall score.   

\section{Limitation}
As we mentioned in the early section, one of the major limitation is the absence of the evaluation of \answerspans~and analytic scores. In future study, we will focus on this evaluation.

\section{Conclusion and Future Study}
We developed automated short answer grading models using only $16$ answers per question, that was less than 5\% of the bench-marking model used. The model was based on the one-shot prompting and Sentence-BERT model. The model achieved a good performance, resulting in 27\% of relative error reduction in accuracy over the majority baseline. 

However, there was a large gap compared to the model trained on the large manually scored dataset. In order to use automated scoring models to grade real tests, the models need to achieve near-human performance, and the performance of the current model was far behind. However, LLM-based prompting can be used in different scenarios. For instance, the analysis of the justification key revealed that the automated justification key extraction was highly accurate. The LLM-based prompting can be used to efficiently and cost-effectively enrich the annotations of the existing dataset, and this dataset can be used to train the models to provide not only the final score, but also additional information that can be used to increase the model's interpretability. Previous studies, such as \newcite{mizumoto2019}, have shown that a joint model trained on predicting both justification keys and holistic scores can improve the accuracy of automated scores.

\bibliography{SAS}
\end{document}